\documentclass[letterpaper, 10 pt, conference]{ieeeconf}  

\IEEEoverridecommandlockouts                              

\usepackage{amsmath,amssymb,mathtools}
\usepackage{bm}
\usepackage{bbm}
\usepackage{graphicx}
\usepackage[hidelinks]{hyperref}
\usepackage{microtype}
\usepackage{todonotes}

\usepackage{xcolor}
\definecolor{myblue}{RGB}{0,90,160}

\allowdisplaybreaks
\title{\LARGE \bf
Learning Quadruped Walking from Seconds of Demonstration
}
\author{
Ruipeng Zhang,
Hongzhan Yu,
Ya-Chien Chang,
Chenghao Li,
Henrik I. Christensen,
Sicun Gao\\[0.4em]
University of California, San Diego\\[0.25em]
\normalsize \textcolor{myblue}{\url{https://latent-variation-regularization.github.io/}}
}

\begin{document}
\maketitle

\thispagestyle{empty}
\pagestyle{empty}

\begin{abstract}
Quadruped locomotion provides a natural setting for understanding when model-free learning can outperform model-based control design, by exploiting data patterns to bypass the difficulty of optimizing over discrete contacts and the combinatorial explosion of mode changes. We give a principled analysis of why imitation learning with quadrupeds can be inherently effective in a small data regime, based on the structure of its limit cycles, Poincar\'e return maps, and local numerical properties of neural networks. The understanding motivates a new imitation learning method that regulates the alignment between variations in a latent space and those over the output actions. Hardware experiments confirm that a few seconds of demonstration is sufficient to train various locomotion policies from scratch entirely offline with reasonable robustness. 
\end{abstract}
\section{Introduction}

Learning-based methods are key to the recent advances in quadruped control, achieving agile behaviors using deep neural network policies~\cite{rudin2022learning, peng2020learning}. Still, a common concern is that training deep neural policies requires a large number of trial interactions with the environment that is only realistic in simulation, and thus their performance suffers from the gap between simulation and hardware~\cite{tang2025deep}. We ask the following question: in a purely offline imitation setting, how much data is actually needed for training deep neural policies for quadruped locomotion from scratch? 

We pose this question specifically for quadruped walking, because it is a natural setting where, intuitively, data-driven methods can have the potential of showing clear advantage over model-based ones. From a control-theoretic perspective, the dynamics of the quadrupeds can be very difficult to handle: it is mainly controlled through contact, while each contact between the legs and the ground incurs discrete events, making it inherently difficult for standard control and optimization methods~\cite{westervelt2018feedback}. Moreover, the four legs generate combinatorial blowup of contact sequences, making it impossible to plan long trajectories without rigid scheduling of the contact mode changes~\cite{posa2016optimization}. On the other hand, from a learning perspective, walking can be enabled by clear regularity in the data pattern of the legs making specific periodic movements. As long as the neural network policies capture such patterns, the multi-leg support should be able to withstand small errors in the movements. This helps explain why four-legged animals achieve stable walking within minutes after birth (unlike human bipedal walking that requires much longer practice and brain development)~\cite{garwicz2009unifying}. Recent advances in learning-based quadruped locomotion approaches are consistent with this view, where neural policies typically exhibit contact patterns that are difficult to schedule manually or analytically but well-captured by data~\cite{dhedin2025simultaneous}.

\begin{figure}
    \centering
    \includegraphics[width=\linewidth]{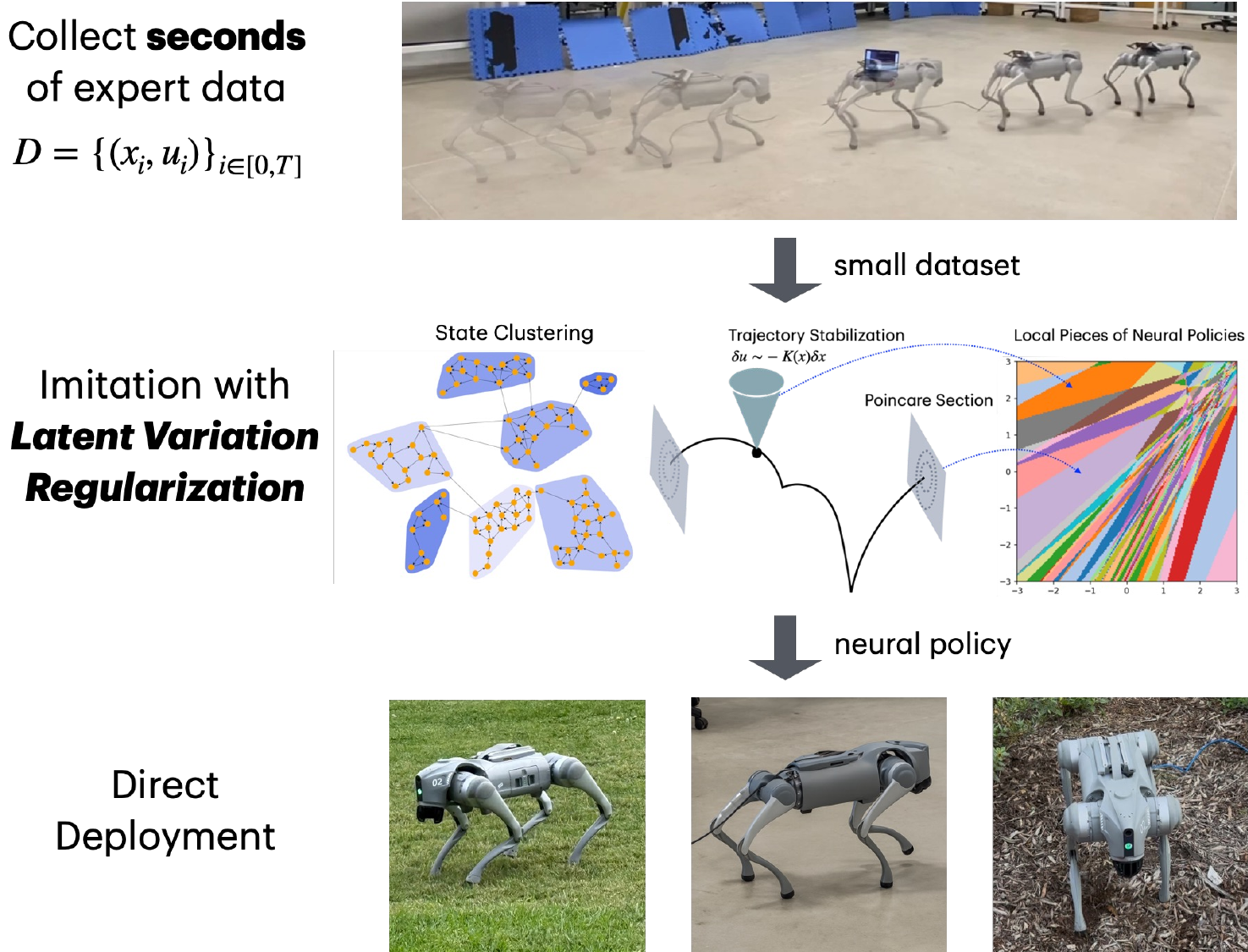}
    \caption{Overview of our approach. We consider the offline imitation learning that collects a small batch of expert demonstration data, and then train deep neural network policies only from the batch without finetuning in simulation or on hardware. We propose new imitation methods with Latent Variation Regularization (LVR) that enforces the matching of the local structure between the control problem and the neural networks. The trained policies are directly tested on hardware platforms with varying ground conditions. }
    \vspace{-0.5cm}
    \label{fig:placeholder}
\end{figure}

In this paper, we analyze how the structure of quadruped walking enables few-sample learning on deep neural networks. We consider the offline setting where the network is trained solely on a fixed batch of expert data without fine-tuning in simulation or on hardware. Our analysis identifies three structural characteristics of this specific control problem:
\begin{itemize}
    \item {\em Local linear structure around stable expert trajectories.} Along a given expert trajectory, the continuous states allow time-varying linearization while the states with discrete jumps can be analyzed through Poincar\'e sections. In both cases, the local stabilization control has an approximately linear structure. 
    \item {\em  Feedforward neural networks allow flexible local fitting.} Given any fixed activation pattern, a deep neural network behaves like a smooth function in a small neighborhood, and the large space of activations decouples these local pieces. Hence local first-order variation constraints can be flexibly enforced over deep neural networks, matching the expressiveness need of the local control maps. 
    \item {\em Sparse critical Poincaré sections.} The walking trajectories are controlled mainly through critical contact events, and the Poincaré sections are visited by a sufficient number of times within a few seconds of demonstration, which enables efficient training of the neural policies. 
\end{itemize}
Based on this understanding, we propose an imitation learning method that enforces the matching of first-order analytic structure in the latent space with the desired local linear feedback control law. The key is to avoid explicit estimation of the linear mapping and convert the matching requirement into minimizing the KL-divergence between the latent variations and the output differentials. This design is critical for bypassing a range of difficulties in the model-free setting with noisy and limited data. 

We conduct hardware experiments to evaluate our analysis and methods. By exploiting the structure and the proposed regularization, deep neural policies trained from a few seconds of demonstration data achieve stable forward, backward, and sideways walking on real quadrupeds. Through testing on varying ground conditions and in simulation, we show that the learning performance is significantly better than behavior cloning that only achieves zero-order fitting without the proposed latent regularization on first-order variations.

\section{Related Work}

Imitation learning for quadruped locomotion with limited demonstration data has been explored along three main directions. First, single- or few-demo pipelines can learn from raw videos to generate reward from observations, followed by training in simulation, enabling successful imitation on simulated quadrupeds \cite{BersethJMLR23} and real robots~\cite{stamatopoulou2024sds}.
Second, motion-priors reduce reward engineering and the amount of robot-collected data by training policies to match reference motions: imitating animal motion produces agile quadruped skills with sim-to-real transfer \cite{PengRSS20} and adversarial motion-prior variants further improve naturalness and robustness in the wild \cite{WangAMPWild2023}.
Third, offline imitation from pre-collected demonstrations scales without on-hardware exploration: diffusion-based controllers trained purely offline reproduce diverse gaits and deploy in real time on quadrupeds \cite{HuangDiffuseLoco2025}, while generalist behavior-cloned policies trained across robots and skills exhibit zero-shot transfer to new platforms \cite{GroqLoco2025}. 

Various lines of existing work have shown that efficient training directly on physical robots typically relies on strong control-theoretic priors that ensure stability during action exploration. The important early work~\cite{tedrake2005learning} showed that a biped can learn to walk in roughly twenty minutes via model-free reinforcement learning with stable backup controller as a prerequisite. In manipulation, \cite{zhao2023learning} employs action-sequence modeling to acquire fine-grained bimanual skills from only minutes of real-world demonstrations. On the model-based control design side, a central design choice in model-based quadruped control is whether the contact sequence is fixed or optimized online.
\cite{ma2020bipedal} achieves rapid gait generation without online mode optimization by reusing fixed gait libraries with prescribed stance-swing timing. Convex MPC approaches typically assume pre-specified schedules and optimize ground-reaction forces with those phases \cite{di2018dynamic}.
Whole-body MPC variants likewise prescribe stance/swing windows via a gait generator and track them with impedance control~\cite{kim2019highly}. To relax hard constraints, 
phase-based trajectory optimization treats contact phases continuously and optimizes timing without integer variables \cite{winkler2018gait}. 

\section{Preliminaries}

\label{app:full_dynamics}

\subsection{Quadruped Dynamics and Hybrid Transitions}


\newcommand{\FL}{\mathrm{FL}}
\newcommand{\FR}{\mathrm{FR}}
\newcommand{\BL}{\mathrm{BL}}
\newcommand{\BR}{\mathrm{BR}}

We consider a quadruped with legs $l\in\{\FL,\FR,\BL,\BR\}$ with three actuated joints (two at the hip, one at the knee). The generalized coordinates are $q\in\mathbb R^{18}$ (6 for the base and 12 joint angles), with velocities $\dot q\in\mathbb R^{18}$, and control inputs $u\in\mathbb R^{12}$. For each foot $l$, the world-frame position is $p_l(q)\in\mathbb R^3$ and the velocity is $\dot p_l(q,\dot q)=J_l(q)\dot q$ where $J_l(q)\in\mathbb R^{3\times 18}$ is the Jacobian. The terrain is the height field $z=\mathcal T(x,y)$ in the world-frame, and the normal gap is
$\phi_l(q) \;=\; e_3^\top p_l(q) - \mathcal T\!\big(e_1^\top p_l(q),\,e_2^\top p_l(q)\big)$.
Write the inertia matrix as $M(q)\in\mathbb R^{18\times 18}$, the Coriolis and centrifugal force $C(q,\dot q)\dot q$, gravity generalized force $g(q)$, and the actuation map $B\in\mathbb R^{18\times 12}$. A contact mode $v$ is specified by a stance set $I_v\subseteq\{\FL,\FR,\BL,\BR\}$. Stacking the stance Jacobians gives $J_v(q)\in\mathbb R^{3|I_v|\times 18}$ and stacking the corresponding contact forces gives $f_c\in\mathbb R^{3|I_v|}$. 

In each contact mode $D_v$, the dynamics is
\begin{equation*}
\label{eq:dae}
\begin{bmatrix} M(q) & -J_v(q)^\top \\[2pt] J_v(q) & 0 \end{bmatrix}
\begin{bmatrix} \ddot q \\[2pt] f_c \end{bmatrix}
=
\begin{bmatrix} B\,u - C(q,\dot q)\dot q - g(q) \\[2pt] -\dot J_v(q,\dot q)\,\dot q \end{bmatrix}
\end{equation*}
subject to constraints that for all $l\in I_v,$ $\phi_l(q)=0$, $J_l(q)\dot q=0$, and for all $j\notin I_v$
$\phi_j(q)\ge h_{\min}$. For each stance foot $l\in I_v$, we write $n_l(q)$ for the terrain unit normal at the current contact point $(x,y)=(e_1^\top p_l(q),e_2^\top p_l(q))$ and $f_l\in\mathbb{R}^3$ for the contact force. With a tangential basis $T_l(q)$ orthogonal to $n_l(q)$, we decompose the contact force as $f_l=T_l(q)\,f_{t,l}+n_l(q)\,f_{n,l}$. Unilateral contact and Coulomb friction impose $f_{n,l}\ge 0$ and $\|f_{t,l}\|_2\le \mu_l f_{n,l}$, where $\mu_l>0$ is the coefficient of friction for foot $l$. The no–slip stance kinematics $J_l(q)\dot q=0$ are physically consistent as long as the inequality $\|f_{t,l}\|_2<\mu_l f_{n,l}$ holds. At impact, the impulse decomposition $\Lambda_l=T_l\Lambda_{t,l}+n_l\Lambda_{n,l}$ must satisfy $\Lambda_{n,l}\ge 0$ and $\|\Lambda_{t,l}\|_2\le \mu_l \Lambda_{n,l}$ so that the reset \eqref{eq:impact_reset} is consistent with friction. Transitions occur at lift-off and impact. A stance foot $l^\ast\in I_v$ lifts off when $f_{n,l^\ast}(q,\dot q,u)=0$ and $\ddot\phi_{l^\ast}(q,\dot q,u)>0$, 
with $q^+=q^-$, $\dot q^+=\dot q^-$, and $I_{v^+}=I_v\setminus\{l^\ast\}$. A swing foot $l^\ast\notin I_v$ impacts when $\phi_{l^\ast}(q)=0$, $\dot\phi_{l^\ast}(q,\dot q)<0$, with reset
\begin{equation*}
\label{eq:impact_reset}
    \begin{bmatrix} M(q) & -J_{v^+}(q)^\top \\ J_{v^+}(q) & 0 \end{bmatrix}
\begin{bmatrix} \dot q^+ \\ \Lambda \end{bmatrix}
=
\begin{bmatrix} M(q)\dot q^- \\ 0 \end{bmatrix}
\end{equation*}
with $q^+=q^-$, $I_{v^+}=I_v\cup\{l^\ast\}$, and $\Lambda\in\mathbb R^{3|I_{v^+}|}$. 

It is clear that the trajectory optimization problem subject to these dynamic constraints has exponential complexity. With four legs and two contact states per leg there are up to $2^4=16$ modes at each instant and grows on the order of $16^k$ over $k$ timesteps. Direct collocation with binary variables for stance versus swing and stick versus slip yields a mixed-integer nonlinear program that is typically intractable. Thus, model-based control design typically enforces fixed model scheduling and carefully handle contact events, and thus generally have difficulty delivering agile maneuvers. 

\subsection{Limit Cycles and Poincar\'e Return Maps}

Stable walking on quadruped corresponds to trajectories that form \emph{limit cycles}, which are periodic solutions satisfying $x^\star(t)=(q^\star(t),\dot q^\star(t))$ with period $T>0$ such that $x^\star(t+T)=x^\star(t)$, and the same sequence of lift–offs and impacts repeats every period. Because of the cyclic nature, we can further replace time by a single \emph{Poincaré section} $\Sigma$, which is a set of states reached once per period, such as when a swing foot $l^\circ$ hits the ground 
$\Sigma \;=\; \{\, x=(q,\dot q):\ \phi_{l^\circ}(q)=0,\ \dot \phi_{l^\circ}(q,\dot q)<0 \,\}$. Starting from $x_k\in\Sigma$, and letting the hybrid flow trajectory returns to $\Sigma$ gives the \emph{Poincaré return map} $P:\Sigma\to\Sigma$, $x_{k+1}\;=\;P(x_k)$. The cycle intersects $\Sigma$ at a fixed point $x^\star\in\Sigma$ with $P(x^\star)=x^\star$. Local orbital stability reduces to the discrete linear system obtained by linearizing $P$ at $x^\star$. The gait is orbitally stable if and only if all eigenvalues of the linearization of $P$ have magnitude strictly less than one~\cite{wiggins2003nonlinear}.

\section{Feasibility of Efficient Learning}

We analyze the structure of the learning problem of quadruped walking. The expert trajectories are hybrid in nature: continuous phases admit trajectory stabilization via local linearization, while discrete contact events define Poincaré sections that summarize step-to-step evolution. In both cases there exist local neighborhoods that admit linear mapping from small state deviations to corrective control. On the other hand, deep neural network policies align naturally with this structure: the high-dimensional parameter space induces a large number of local pieces that are approximately independent from each other, so that each local piece can be trained to match the needed stabilizing slope around different states along the trajectory. Moreover, when the data covers neighborhoods of the critical states, especially the section states immediately before and after contacts, then the local pieces of the neural policy can match the stabilizing feedback along continuous arcs and across discrete events. Thus their composition generates an approximately contracting return map that enables stable walking.

\subsection{Local Stabilizability around Expert Trajectories}

Let $\tau$ be an expert trajectory with $\tau(t) = (\hat x(t),\hat u(t))$ and $t\in \{0,...,T\}$. Assuming that the expert trajectory represents stable walking and is a limit cycle, we can focus on the case where $t=0$ and $t=T$ are two visits at a Poincaré section $\Sigma$, which means $\tau$ is exactly one period of the limit cycle. Because of the hybrid nature of the stable walking trajectory, there are two types of states among $\tau(t)$. The continuous component $\tau_c$ consists of all states where the Jacobian of the system dynamics exists. The discrete jump component $\tau_d$ consists of states that incur discrete jumps in some dimensions in the system dynamics because of contact events. We analyze local stabilization around the expert trajectory in both cases. 

Around an arbitrary state $\hat x(t_i)$ in the continuous component $\tau_c$, the Jacobian of the dynamics exists, which means for states $x(t)$ in a small neighborhood around $\hat x(t_i)$, the variations on the state and control 
\[
\delta x(t)=x(t)-\hat x(t_i),\qquad \delta u(t)=u(t)-\hat u(t_i).
\]
follow the local time-varying linear dynamics
\[
\dot{\delta x}(t)=A_i(t)\,\delta x(t)+B_i(t)\,\delta u(t),
\]
Thus the local stabilization problem around the desired equilibrium $\delta x=0$ is an LQR problem:
\[
\min_{\delta u} \int_{t_i}^{t_{i+1}}\!\big(\delta x(t)^\top Q_i(t)\,\delta x(t)+\delta u(t)^\top R_i(t)\,\delta u(t)\big)\,dt.
\]
with positive semidefinite costs $Q_i(t)$ and $R_i(t)$. The optimal feedback is thus a linear local control law
\[
\delta u(t)=-K_i(t)\,\delta x(t),\qquad K_i(t)=R_i(t)^{-1}B_i(t)^\top P_i(t),
\]
where $P_i(t)$ solves the finite-horizon Riccati equation. Because of the optimality of this local control, we expect that around the expert trajectory that exhibits stable walking, the mapping from state variation to action variation should be approximately linear (without requiring analytic knowledge of what the gain matrix needs to be). 

On states in the discrete jump component $\tau_d$ of the expert trajectory, the instantaneous Jacobian is undefined, so the trajectory stabilization formulation above does not apply. We can still analyze these states rigorously by considering the Poincaré sections that they are at. Let $\Sigma$ be such a Poincar\'e section and let $\hat x_\Sigma\in\Sigma$ be an approximate fixed point on the expert trajectory. Assuming that the expert achieves stable walking, there exists a Poincar\'e return map such that $P(\hat x_\Sigma)=\hat x_\Sigma$ (approximately in practice, writing equality for simplicity). For a small variation $\xi_k=x_k-\hat x_\Sigma$ on $\Sigma$, we can now consider the Jacobian $A_P$ of the return map $P$ so that the next intersection satisfies $\xi_{k+1}\;=\;A_P\xi_k$ (under appropriate control as part of $P$),
which is again a linear approximation near the fixed point. Hence local stability on the section can be checked by the eigenvalues of $A_P$, and a stabilizing control law near the section is also well approximated by a linear feedback in the state error. To connect this analysis with local control, let $t_j\in(0,T)$ be the contact time. Use two tiny windows that straddle $t_j$. On the pre-impact window $[t_j-\varepsilon,t_j)$ apply local stabilization
\[
\delta u(t)=-K_j^-\,\delta x(t),\qquad \delta x(t_j^-)=\Psi_j^-\,\delta x(t_j-\varepsilon),
\]
and on the post-impact window $(t_j,t_j+\varepsilon]$ apply
\[
\delta u(t)=-K_j^+\,\delta x(t),\qquad \delta x(t_j+\varepsilon)=\Psi_j^+\,\delta x(t_j^+).
\]
writing the jump as the reset $x^+=R_j(x^-)$ and linearizing:
\[
\delta x(t_j^+)=R_j\,\delta x(t_j^-).
\]
Consequently, combining the two windows and the reset,
\[
\delta x(t_j+\varepsilon)=\Psi_j^+\,R_j\,\Psi_j^-\,\delta x(t_j-\varepsilon).
\]
Over one period, composing all such pre/post windows and smooth segments, and then projecting to $\Sigma$, yields the return Jacobian $A_P$. Because $A_P$ is linear in the local deviations, choosing the pre/post laws to be linear makes the resulting step-to-step map near $\hat x_\Sigma$ linear as well. If all eigenvalues of $A_P$ have magnitude less than one, then errors on the section contract and the gait is stable.

A special structure of the quadruped walking motion is that in the periodic gaits with a fixed event order, only a small subset of the Poincar\'e sections are critical, which are sparse sections that anchor the limit cycle. The product of closed-loop transitions between these anchors yields the same return matrix as any finer partition, so enforcing accurate local linear feedback on the anchors is sufficient. For instance, swing-leg states away from anchors need not match exactly each cycle; as long as the anchor states are matched, their deviations do not change the step-to-step map and do not affect the stability of the overall behavior. Consequently, we do not need to match and stabilize every state along the expert trajectory, but can focus on the critical states. Indeed, in the next subsection, we consider the local numerical properties of deep neural networks to show that it is likely that the data collected over a short duration of expert demonstration is enough to pin down the local feedback needed for stable walking. 

\subsection{Local Approximation with Neural Networks}

Consider a deep network $\pi(x,\theta)$ with common activations. Around an input state $\hat x_i$ (for a fixed phase $\theta_i$), there is a small neighborhood where the activation pattern does not change. In such neighborhood with fixed activation units, the network is reduced to a smooth function~\cite{yu2024activation}.  
For instance, if the activation is ReLU, these neighborhoods are polyhedral regions that are natural templates for fitting the local feedback stabilization control laws. Moreover, because different activation patterns use different activations, the subset of parameters that are activated in one region can have limited overlap with those in other regions, which produces near-independence across pieces. When the target local law at some neighborhood is linear, matching the local stabilization amounts to minimizing the usual squared error loss over the corrective control variation, so even standard training on the neural network can behave like linear regression locally and converge to the best approximately affine fit that matches the local feedback needed. 

In a small neighborhood $U_i$ around an anchor state $\hat x_i$, a neural network with common activations behaves like an affine map, so learning the local law $\delta u=-K_i\,\delta x$ reduces to estimating a linear gain. Crucially, the data needed does not depend on the raw number of network parameters: once the activation pattern is fixed on $U_i$, training with squared error is equivalent to linear regression for $K_i$. Let $p$ be the number of free coefficients in $K_i$. If the samples $(\hat x_j,\hat u_j)$ in $U_i$ are well spread with $\lambda_{\min}\!\Big(\mathbb{E}\big[(\hat x-\hat x_i)(\hat x-\hat x_i)^\top\big]\Big)\ \ge\ \lambda_0>0$,
then standard least-squares bounds imply that, with probability at least $1-\delta$,
$\sup_{\|x-\hat x_i\|\le r_i}\ \big\|(\widehat K_i-K_i)(x-\hat x_i)\big\|\ \le\ \varepsilon$
when the number of samples reaches
$N_i\ \ge\ C (p+\log\tfrac{1}{\delta})(\lambda_0\,\varepsilon^2)^{-1}$ for some universal constant $C$ \cite{Wainwright2019}. Thus, the local sample complexity scales with the dimension $p$ of the feedback law, not with the size of the neural network. The total samples across $M$ anchors are $\sum_{i=1}^M N_i$, growing linearly in $M$ and as $\varepsilon^{-2}$ in the desired accuracy.

\begin{figure}
    \centering
    \includegraphics[width=\linewidth]{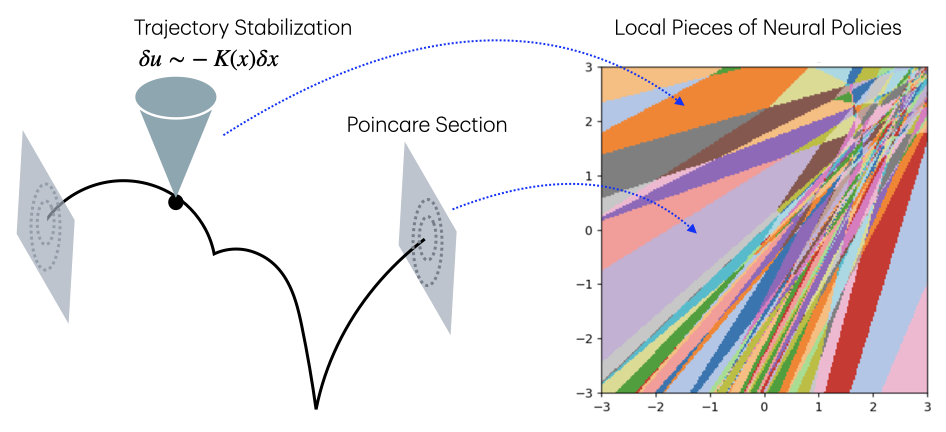}
    \caption{Illustration of the first-order variation requirement in neural policies. Around expert trajectories of stable walking, the local stabilizing control laws are linear at both continuous (analyzed through trajectory stabilization) and discrete jump states (analyzed through Poincar\'e sections). This structure matches with the local smooth pieces in deep neural networks that are approximately independent because of sparsity in the large parameter space. Thus the local feedback requirements can be readily enforced through imitation learning that regularize local variations in the latent space.}
    \label{fig:placeholder}
    \vspace{-.3cm}
\end{figure}

\section{Imitation by Latent Variation Regularization}

Given the analysis in the previous section, it should be clear that training stable quadruped walking by imitation from expert trajectories is feasible in a small-data regime. However, it is still unclear how to algorithmically exploit the structure in a model-free setting, because:
\begin{itemize}
\item The analysis makes frequent use of optimal local control laws and idealize that the expert trajectories produce limit cycles. In practice, the data collected from demonstrations will be noisy. Since we do not assume any prior knowledge of the models, calculating the ideal local stabilization gains at each state is infeasible in a small-data regime. 
\item We hypothesized that the critical states for stabilizing the trajectories should be sparse, and for an expert trajectory with reasonable length, the periodic samples should have enough coverage of the neighborhoods of the critical states. However, if we only perform standard behavior cloning there is no easy analysis on the samples for sufficient coverage. 
\item If we use a standard mean squared error loss on matching the network output with data (which is behavior cloning), then the local piecewise structure of the network may not be utilized. It is likely that all inputs are projected into sparse locations in the latent space, and the slope in their local neighborhood cannot be matched with the desired local control feedback through data. 
\end{itemize}
Consequently, we now propose a specific design for imitation learning, still in the offline setting training from small dataset from scratch, that can exploit the structure of the problem through regularizing the latent representation in the network to efficiently approximate locally stable control policies. 

\subsection{Neural Features for Local Stabilization}

We aim to train deep neural network policies $\pi_{\theta}: X\rightarrow U$ whose local properties match the time-varying local stabilization structure as analyzed above. We focus on the standard feedforward multilayer perceptron architecture (MLP). Although such neural networks are highly nonlinear functions, the input of the last layer always goes through a linear transformation into the output. Now if we consider a small variation on the input state $x+\delta x$, locally expanding the network gives us a mapping of the form
\[
\delta u \;=\; W\cdot D_x\phi_\theta(x, \delta x),
\]
where $D_x\phi_{\theta}$ is the Jacobian (or a suitable approximation at non-smooth points) over the inputs $x$ of the latent mapping $\phi_{\theta}$ of the original policy network at the specific input state, where $\theta, W$ are parameters of the original policy network $\pi_\theta$. Note that the bias on the last layer disappears in this mapping after differentiation. 

Consequently, we can view $\phi_{\theta}$ as a feature mapping that captures {\em latent variation} $\delta h = D_x\phi_\theta(x, \delta x)$ at the latent vector $h$ of the original input $x$. If we can match $W$ with the time-varying gain $K_i(x,t)$ discussed in the previous section, then we can have a perfect feedback local control law for stabilizing the trajectories. However, this is impossible, because $W$ is not dependent on the input $x$ or time $t$ (training time-varying control policy will require significantly more data that is not considered in our setting). Moreover, as mentioned above, we do not assume model-based understanding of $K_i(x,t)$ or enough data access to estimate it across all data points. 

Our main focus of the algorithm design is to bypass this difficulty. The key is to treat $\phi_\theta$ as a kernel function so that 
\[\mathrm{dist}(D_x\phi_\theta(x,\delta x), D_x\phi_\theta(x',\delta x')) \mbox{ and } \mathrm{dist}(\delta u, \delta u')\]
match under a suitable distance definition $\mathrm{dist}(\cdot, \cdot)$, which is the case when $\delta x$ and $\delta u$ has the linear connection in ground truth. In this way, we avoid estimating the gain matrix between $\delta x$ and $\delta u$ and also will be able to work with a state-independent $W$ by incorporating state-varying factors into the feature mapping $\phi_{\theta}(x,\delta x)$ itself. This is the core of our algorithm design described in the next part. Importantly, we turn the matching conditions into distributional forms that can be applied over the entire dataset uniformly.  

\subsection{Imitation Algorithm and Regularizer Design}

Standard behavior cloning makes use of the dataset by imposing zero-order fit of the state-to-action map in the expert trajectory, typically through MSE or cross-entropy loss. As shown in the previous sections, stable walking in closed-loop is critically governed by reliable local feedback control laws that require {\em first-order} fit to the {\em variations} in state and control from the expert trajectory. We achieve this by adding the following operations on data. 

First, from the collected dataset, we construct a $k$-Nearest-Neighboring (KNN) graph. The primary goal is to identify critical variations in the states to enable the learning of local structures. Let $\{x_{i}, u^{*}_{i}\}_{i=1}^{T}$ be expert demonstration samples. We construct a kNN graph $G = (V, E)$ with $V=\{x_1,...,x_T\}$ and $(x_i, x_j) \in E$ if $x_{j}$ is among the $k$ nearest neighbors of $x_{i}$ under Euclidean distance in state space.
To enforce strict locality, 
we prune edges by a node-wise radius:
compute $\varepsilon_{i}$ as the $q$-quantile of $\{d(x_i,x_j): (i,j)\in E\}$ and retain only edges with $d(x_{i}, x_{j}) < \varepsilon_{i}$.

Then, starting from a randomly initialized candidate neural network $\pi_\theta$, we map all inputs to a latent space $\mathcal{H}$ via the feature map $\phi_{\theta}$ that preserves the graph structure defined by edges $e=(i, j)\in E$. For each edge, the latent chord $\delta h_e =\phi_{\theta}(x_j)-\phi_{\theta}(x_i)$ satisfies:
\[
    \delta h_e \approx D_x\phi_{\theta}(x_i)\,(x_j-x_i),
\]
where $x_j-x_i$ is considered as $\delta x$ and thus capturing the local tangent structure in $\mathcal{H}$.
In control space, the corresponding expert increment $
\delta u = u_j- u_i$ is the local control response.

Now for each edge $e$ in $\mathcal{H}(G)$,
we collect a neighborhood $N(e) \subseteq \mathcal{H}(G)$.
Define $\delta h_{e,\parallel} := P_{\mathrm{row}}\delta h_e$, where $P_{\mathrm{row}} = W^\top(WW^\top)^{+}W$ and $(\cdot)^{+}$ denotes the Moore–Penrose pseudo-inverse. This projects $\delta h_e$ onto the control-relevant row space of $W$, filtering out variations in control-invariant directions. We then define the local orientation distribution as the softmax distribution over the cosine distance among the variations:
\begin{equation}
    p_{\mathcal{H}}(j|e) = \frac{\exp(\cos(\delta h_{e, \parallel}, \delta h_{j, \parallel})/\tau)}{\sum_k \exp(\cos(\delta h_{e, \parallel}, \delta h_{k, \parallel})/\tau)}.
\end{equation}
where $\tau > 0$ is a temperature parameter. The corresponding control space orientation distribution is:
\begin{equation}
    p_{\mathcal{U}}(j|e) = \frac{\exp(\cos(\delta u_{e}^*, \delta u^*_j)/\tau)}{\sum_k \exp(\cos(\delta u_{k}^*, \delta u^*_j)/\tau)}.
\end{equation}
We then align these local orientation distributions by minimizing their KL-divergence:
\begin{equation}
\label{eq:overall-loss}
    \mathcal{L}_{\mathrm{KL}}
=\mathbb{E}_{e \in \mathcal{H}(G)}[ \mathrm{KL}\!\big(p_{\mathcal{H}}(\cdot\mid e)\;\|\;p_{\mathcal{U}}(\cdot\mid e)\big)].
\end{equation}
Write the standard behavior cloning objective as
\[
\mathcal{L}_{\mathrm{BC}}=\mathbb{E}_i\big\|W\,\phi_{\theta}(s_i)+b-u^{*}_i\big\|^{2}_{2},
\]
which ensures zero-order fit of the mapping. 
Combining, we optimize the overall loss:
\begin{equation}
\mathcal{L}=\mathcal{L}_{\mathrm{BC}}+\lambda\,\mathcal{L}_{\mathrm{KL}},
\end{equation}
with non-negative weight $\lambda$.
The KL term enforces that small, local moves in the control-relevant latent directions induce consistency oriented action changes,
improving interpolation and robustness to modest distribution shift.

\begin{figure*}
    \centering
    \includegraphics[width=0.9\linewidth]{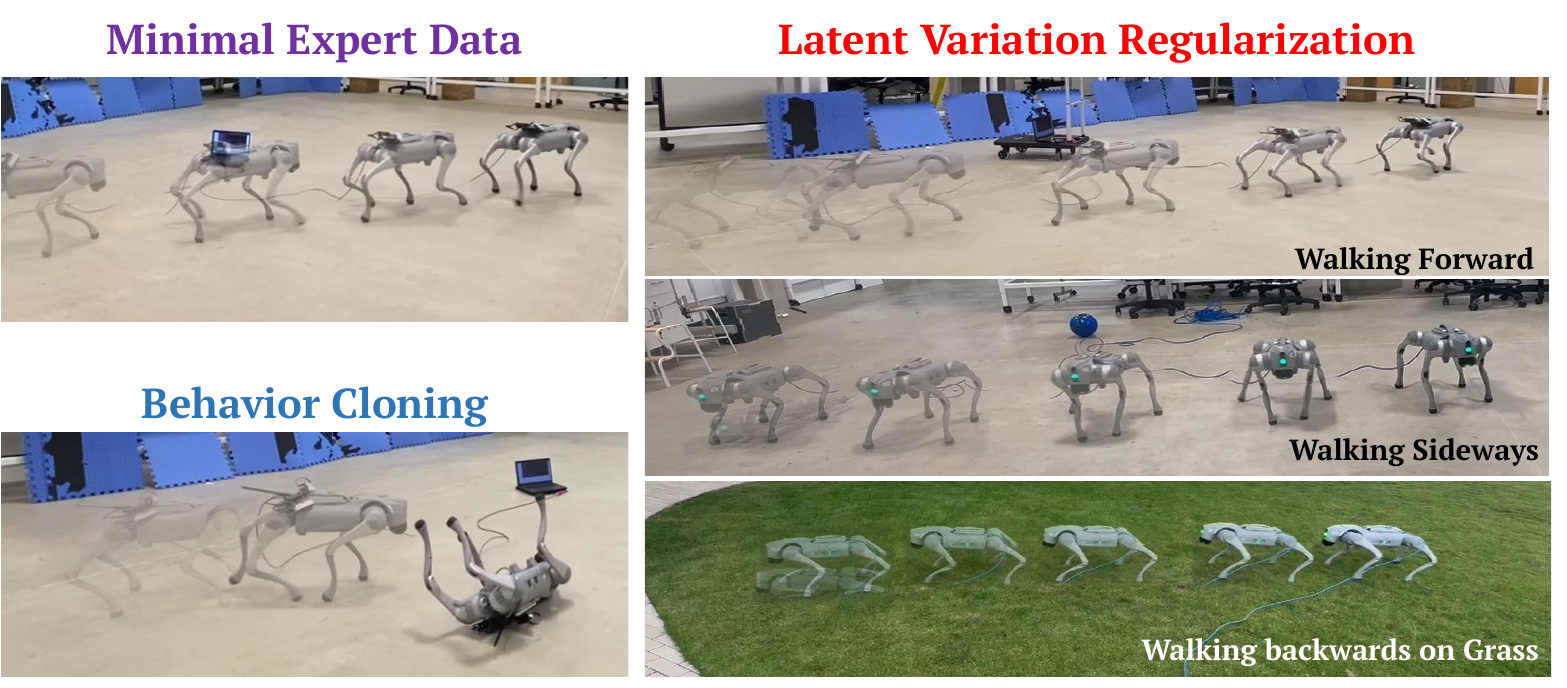}
    \caption{Real-world deployment of quadruped policies trained with minimal expert data. The upper left shows expert demonstration. Using the same data, behavior cloning (bottom left) fails to walk, whereas our Latent Variation Regularization (LVR, right column) produces stable forward and sideways walking. LVR further shows robust performance, such as walking backwards on grass by training on data from walking backwards on flat indoor ground.}
    \label{fig:realworld}
\end{figure*}

\section{Experiments and Analysis}

In this section, we first highlight the data efficiency of our approach, demonstrating that Latent Variation Regularization (LVR) enables strong performance with limited data. We then analyze the learned latent space to better understand the effect of the LVR loss. Next, we evaluate the robustness of the resulting policies in simulation, and finally present real-world deployment results.

\noindent\textbf{Experiment Setup.} We conduct experiments on the Unitree Go2 quadruped and its corresponding IsaacLab simulator. 
Expert RL policies are used to collect demonstration trajectories for imitation learning in the simulator and on the real robot over flat ground in the lab.
All experiments use MLP neural network policies with 3 hidden layers of 128 neurons each, and ELU as activation function. We set $K=32$ nearest neighbors and $\tau,\lambda=0.1$ across experiments.

\begin{figure*}[h!]
    \centering
    \includegraphics[width=0.9\linewidth]{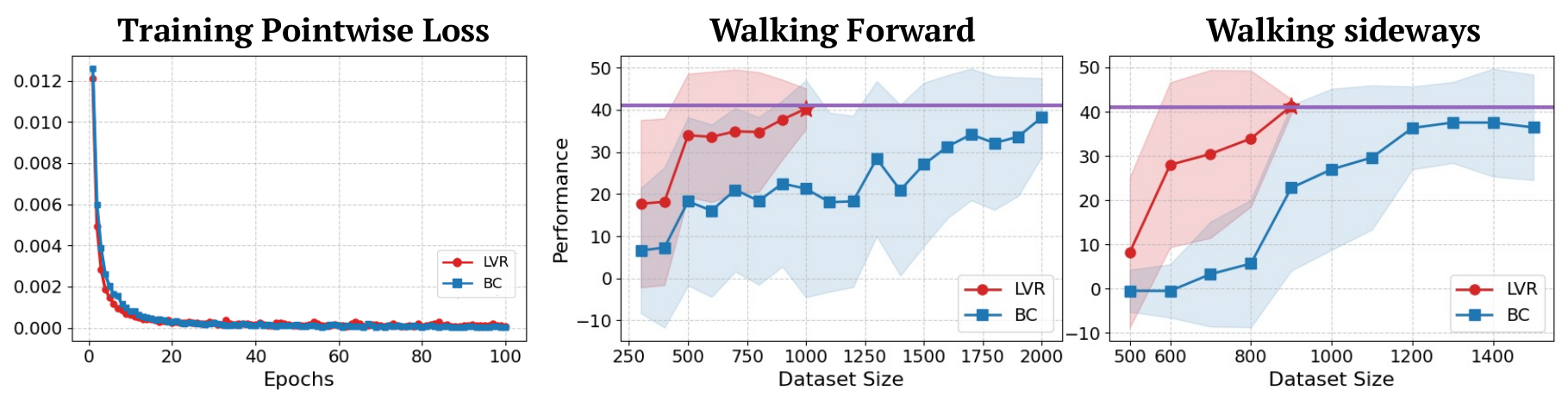}
    \caption{Left: training pointwise imitation loss, where both BC and LVR rapidly converge to similar values. Right: control performance on forward and sideways walking with varying dataset sizes. LVR achieves expert-level performance with $\leq$ 1 trajectory, whereas BC requires substantially more demonstrations to approach similar returns. Purple line denotes the expert policy performance.}
    \label{fig:data-perf}
\end{figure*}

\subsection{Data Efficiency}

The key advantage of matching latent variation is its high sample efficiency. In Figure~\ref{fig:data-perf}, we compare LVR with behavior cloning across varying amounts of demonstration data. For each dataset size, policies are trained with supervised imitation learning for a fixed number of epochs and then evaluated in simulation over 100 rollouts, and the plots report the mean and standard deviation.

The results show that LVR consistently achieves expert-level performance within one trajectory, whereas behavior cloning requires much larger datasets to approach similar returns. Notably, despite both methods rapidly minimizing the pointwise imitation loss, only LVR translates this into robust control policies that generalize effectively from limited data. This demonstrates that pointwise matching alone is insufficient to guarantee strong control performance, and that enforcing structure in the latent variation dynamics is crucial for sample-efficient imitation learning. These findings highlight the data efficiency of LVR and its potential to make imitation learning feasible in domains where expert data is costly or difficult to collect.

\subsection{Latent Feature Space Analysis}
\label{sec:latent}

\begin{figure}
    \centering
    \includegraphics[width=1\linewidth]{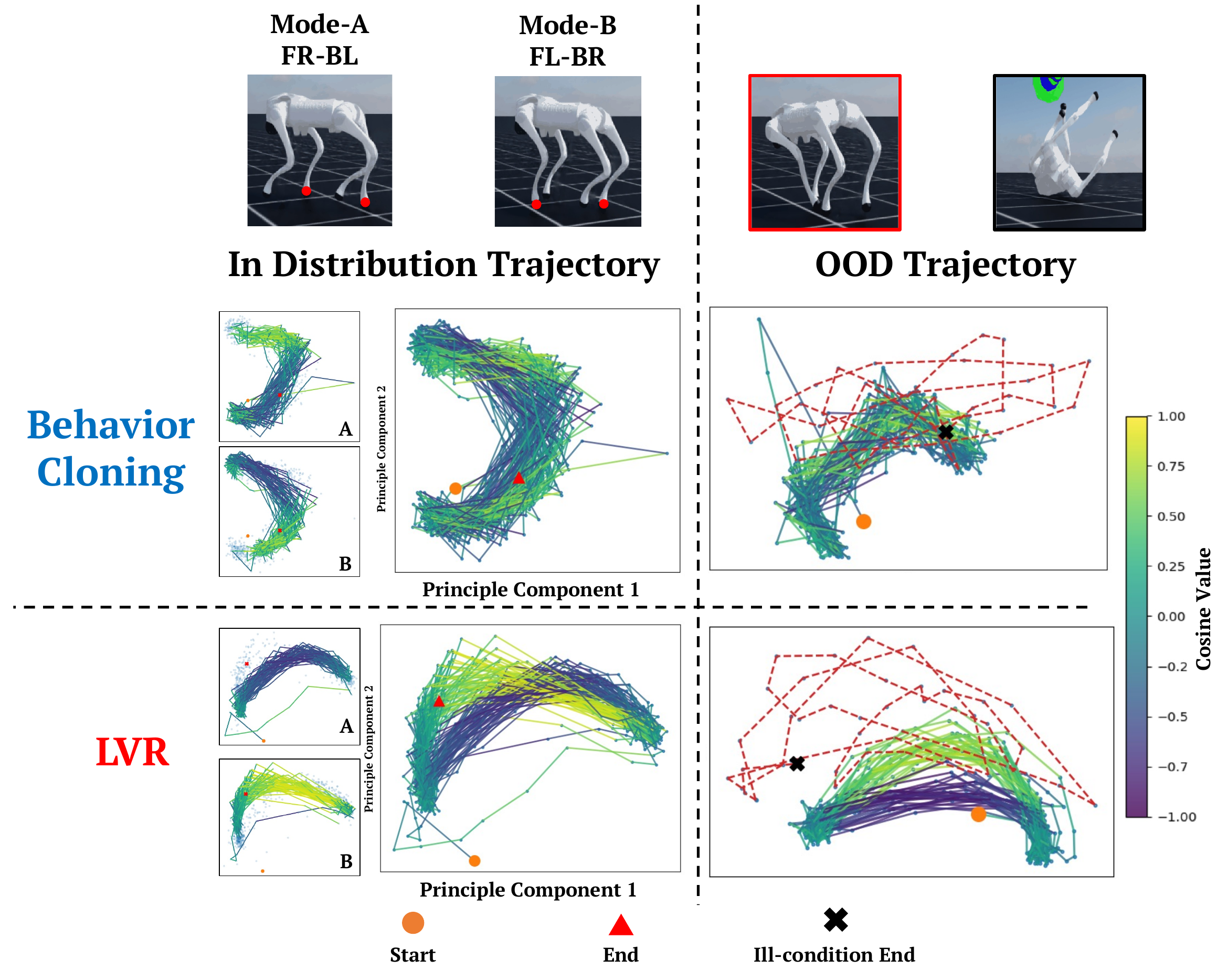}
    \caption{PCA visualization of latent states $h_t$, where projection axes are obtained from PCA on consecutive differences $\delta h_t=h_{t+1}-h_t$. Points are connected in temporal order and colored by cosine similarity of $\delta h_t$
with the first principal component (PC1). Left: expert forward-walking trajectory showing structured trot dynamics with two alternating gait modes. Right: an OOD rollout where BC (top) collapses while LVR (bottom) preserves coherent orientation bundles and separates ill-conditioned late states. See details in Section~\ref{sec:latent}.
 }
    \label{fig:hidden-traj}
    \vspace{-.6cm}
\end{figure}

Our method enforces alignment between the orientation distribution of latent feature differences and the corresponding oracle control differences.
The key idea is that if this alignment succeeds, the latent representation will not only reproduce the expert’s control mappings but also preserve the local variation structure induced by locomotion dynamics. Since expert controls evolve periodically, the sequence of latent states $\{h_t\}$ are expected to trace out a structured trajectory in the latent space $\mathcal{H}$, 
where consecutive differences $\delta h_t=h_{t+1}-h_t$ reflect the geometry of periodic control signals. To examine this structure, we perform principal component analysis (PCA) on the set of $\delta h_t$, and then project the latent states $h_t$ onto the first two principal components. Each point in the plot corresponds to a latent state, with consecutive points connected according to temporal order (Figure~\ref{fig:hidden-traj}).

In the in-distribution case, both behavior cloning and LVR show structured latent state transitions in the walking period. The expert data correspond to the characteristic trot gait, where diagonal pairs of legs alternate between stance and swing phases. When separated into Mode-A (front right and back left legs on the ground) and Mode-B (front left and back right legs on the ground), the corresponding $\delta h_t$ cluster into two bundles, each exhibiting consistent geometry. PCA analysis 
shows that LVR captures a dominant shared direction in latent variation, while BC distributes energy across multiple components. 
This difference suggests that the zero-order imitation loss in BC fails to enforce coherent local linear structure in the latent space.

The implications of this gap become clear in an out-of-distribution rollout in which a BC policy fails, as shown in Figure~\ref{fig:hidden-traj} (right).
We collect the states along this failed trajectory, and feed them into both the BC and LVR policies. 
Before the failure, the latent trajectory produced by LVR continues to exhibit the same linear structure observed in the expert data, whereas BC produces disorganized latent transitions: the local orientations of $\delta h_t$ are inconsistent, and ill-conditioned late states are mapped into the same latent region as earlier healthy states. By contrast, LVR maintains coherent orientation bundles corresponding to the two gait modes and clearly separates OOD states and motions that violate local linearity. 
This analysis highlights that BC’s inability to preserve consistent latent orientations renders it fragile under OOD conditions, while LVR’s preservation of locally linear dynamics enables more reliable detection and handling of distributional shifts.

\begin{figure}
    \centering
    \includegraphics[width=1\linewidth]{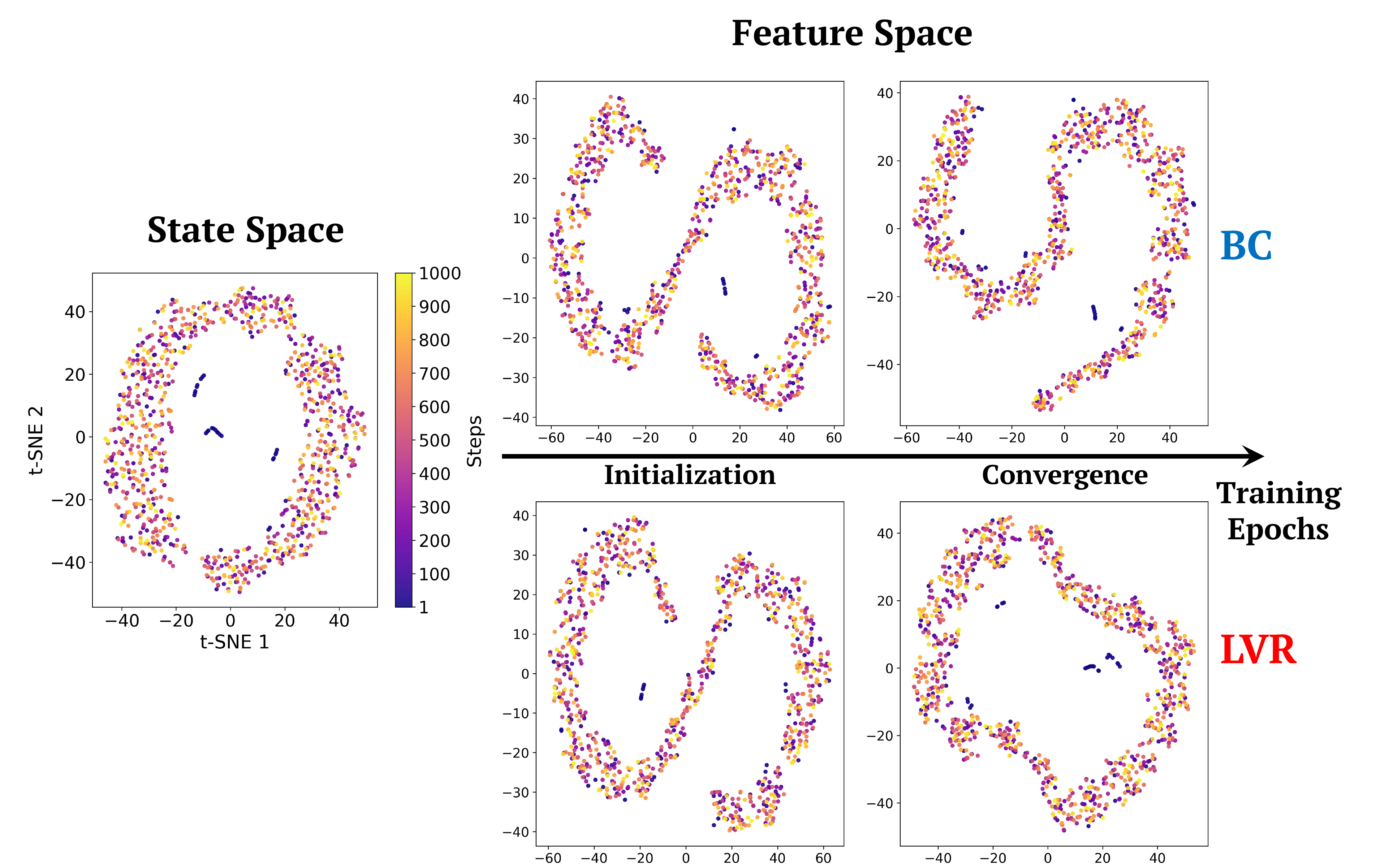}
    \caption{t-SNE visualization (perplexity 15) of the expert training trajectory in state space (left) and the corresponding latent feature spaces learned by BC (top) and LVR (bottom). Points are colored by timestep index in the original trajectory. For the feature-space plots, the left column shows embeddings near initialization, while the right column shows embeddings at convergence. BC embeddings remain fragmented throughout training, whereas LVR progressively organizes local neighborhoods into a coherent loop structure that mirrors the periodic gait cycle observed in the state space.}
    \label{fig:tsne}\vspace{-.2cm}
\end{figure}

To further analyze the geometry of the learned latent space, we apply t-SNE with perplexity 15 to emphasize local neighborhoods. The left panel of Figure~\ref{fig:tsne} shows the state inputs along the expert trajectory, where the cyclic gait appears as a loop-like structure colored by timestep. The right panels visualize the corresponding latent embeddings from BC (top) and LVR (bottom) at two stages of training: initialization (left) and upon convergence (right). Although both methods initially preserve local neighborhoods, their training dynamics diverge. BC embeddings remain fragmented, forming disconnected clusters that fail to recover the global periodic cycle evident in the state space. In contrast, LVR progressively organizes these neighborhoods into a coherent loop structure, closely mirroring the underlying gait dynamics. These results imply that LVR maintains local similarity and tends to produce a more coherent global manifold, which helps improve performance over vanilla behavior cloning.

\subsection{Evaluation on Robustness}
\begin{figure}
    \centering
    \includegraphics[width=1\linewidth]{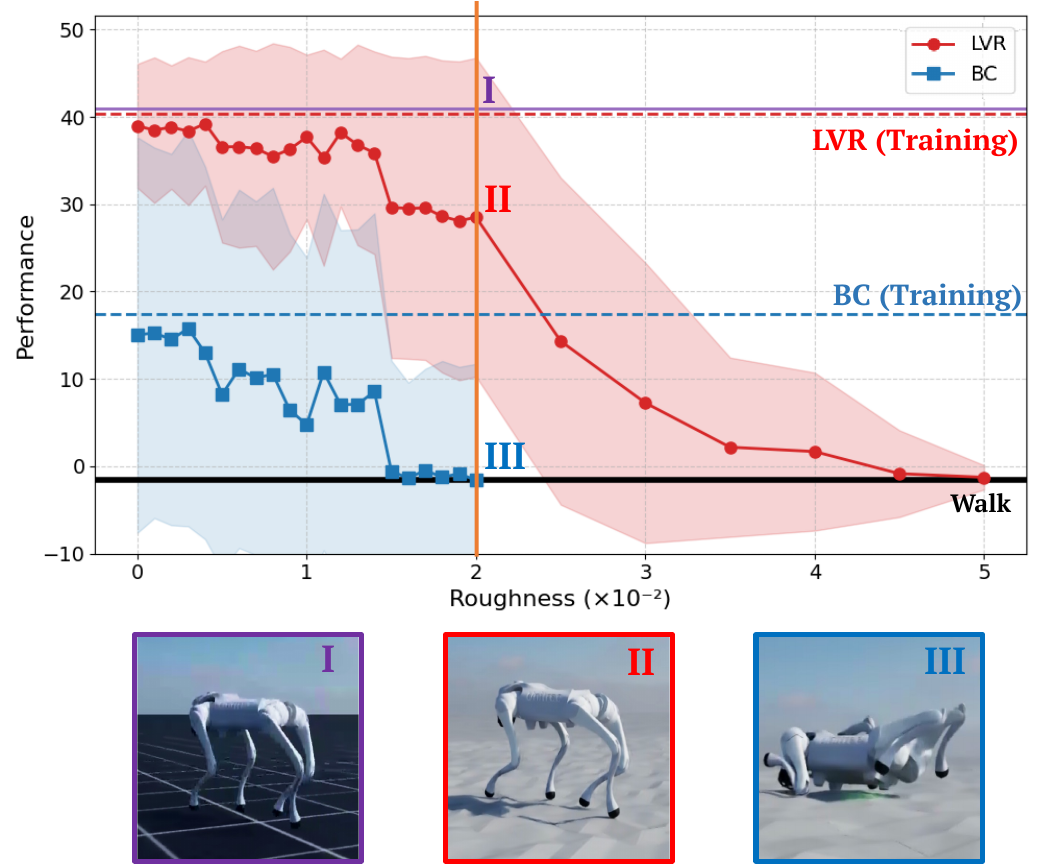}
    \caption{Robustness evaluation of forward walking under increasing terrain roughness. Both BC and LVR are trained on the same demonstrations collected on flat ground and then evaluated on terrains with varying roughness scales. The purple line indicates expert performance on flat terrain, and the black line marks the walkability threshold below which locomotion fails. Rendered images show representative behaviors of the expert (I), LVR (II), and BC (III) policies. LVR sustains high performance across a wider range of roughness levels, whereas BC rapidly deteriorates once the environment deviates from the training condition.}
    \label{fig:robustness}
\end{figure}

Beyond data efficiency, we evaluate robustness to deployment shift by testing policies trained on a single expert trajectory collected on flat terrain on the forward walking task, across environments with increasing terrain roughness. We train both LVR and BC on the same data trajectory and evaluate them under randomized seeds and stochastic perturbation noise. As shown in Figure~\ref{fig:robustness}, LVR sustains high performance across a wide range of roughness levels, whereas BC degrades rapidly once the environment departs from the training condition. Notably, LVR continues to walk in regimes where BC fails entirely. The rendered snapshots illustrate these behaviors (I: expert on flat, II: LVR stable on rough terrain, III: BC collapse).
This robustness stems from LVR’s inductive bias: by aligning the orientation distribution of local latent differences $\delta h$ with the oracle control differences $\delta u$, LVR preserves a consistent local linear structure in $\mathcal{H}$ so that the latent trajectory remains organized and the readout remains predictable even for off-manifold states.

\subsection{Real-world Experiments}

We evaluate LVR on a real Unitree Go2 quadruped to assess its feasibility with real data and its ability to bridge the sim-to-real gap. Tasks include forward walking at different target speeds (0.5 m/s, 0.7 m/s, 1.0 m/s), sideways walking (0.5 m/s, 0.7 m/s), and backwards walking (0.7 m/s). Expert trajectories are 
collected on flat indoor ground. For each task, we collect up to two trajectories consisting of 250 datapoints each (5 seconds at 50 Hz). Policies are trained solely on this data. Tests are conducted on flat ground as well as on more challenging surfaces such as bricks and grass (Figure~\ref{fig:realworld}).
These real-world experiments demonstrate that LVR not only transfers successfully from simulation but also provides robustness in practical deployment conditions after training with minimal expert data.

\section{Discussion and Conclusion}

We studied why quadruped walking can be learned offline from a small batch of demonstrations. We showed that local feedback along an expert steps is well described by local linear feedback while the feedforward neural policies admit locally pieces that can match such requirements on the first-order variations. Building on this understanding, we introduced an imitation method based on latent variation regularization that encourages the network to match first-order behavior without explicitly estimating stabilization gains. We performed extensive experiments and observed that stable walking with deep neural policies can be achieved with imitation from a few seconds of demonstration data. There are many opportunities for future work. A detailed analysis of the learning dynamics under the proposed latent variation regularization would enhance the understanding of the generality of the approach. Combining the proposed offline approach with targeted online data collection and testing on broader platforms and tasks will further improve robustness and generalization of the proposed methods. 

\bibliographystyle{IEEEtran}
\bibliography{refs}

\end{document}